%% file: main.tex
\documentclass[10pt,twocolumn,letterpaper]{article}

\usepackage{iccv}              
\usepackage{graphicx}
\usepackage{amsmath}
\usepackage{amssymb}
\usepackage{enumitem}
\usepackage{bbm}
\usepackage{setspace}
\usepackage{tikz-imagelabels}
\usepackage{pifont}
\usepackage{listings}
\usepackage{subcaption}
\usepackage{array}
\usepackage{arydshln} 
\usepackage{hhline}
\usepackage{booktabs}
\usepackage{multirow}
\usepackage{multicol}
\usepackage{float}     
\usepackage{afterpage} 


\definecolor{iccvblue}{rgb}{0.21,0.49,0.74}
\usepackage[pagebackref,breaklinks,colorlinks,allcolors=iccvblue]{hyperref}

\def\paperID{8928} 
\def\confName{ICCV}
\def\confYear{2025}

\title{Knowledge Consultation for Semi-Supervised Semantic Segmentation}

\author{
Thuan Than$^1$ \hspace{0.1cm} Nhat-Anh Nguyen-Dang$^1$ \hspace{0.1cm} Dung Nguyen$^1$ \hspace{0.1cm} Salwa K. Al Khatib$^2$ \hspace{0.1cm} Ahmed Elhagry$^2$ \\
Hai Phan$^3$ \hspace{0.1cm} Yihui He$^3$ \hspace{0.1cm} Zhiqiang Shen$^2$ \hspace{0.1cm} Marios Savvides$^3$ \hspace{0.1cm} Dang Huynh$^1$ \\
\tt\small
\{thandoanthuan, pthai1204, huynhthedang, zhiqiangshen0214\}@gmail.com \\
\tt\small
\{anh.nguyen.220069, dung.nguyen.190014\}@student.fulbright.edu.vn \\
\tt\small
\{salwa.khatib, ahmed.elhagry\}@mbzuai.ac.ae \hspace{0.5cm}
\tt\small
he2@alumni.cmu.edu \hspace{0.5cm}
marioss@andrew.cmu.edu \\
$^1$Fulbright University Vietnam \quad
$^2$Mohamed bin Zayed University of AI \quad
$^3$Carnegie Mellon University}

\begin{document}
\maketitle

\input{section/0_abstract}    
\input{section/1_introduction}
\input{section/2_related_work}

\input{section/3_method}

\input{section/4_experiments}
\input{section/5_conclusion}
{
    \small
    \bibliographystyle{ieeenat_fullname}
    \bibliography{main}
}

\end{document}

%% file: section/0_abstract.tex
\begin{abstract}
Semi-Supervised Semantic Segmentation reduces reliance on extensive annotations by using unlabeled data and state-of-the-art models to improve overall performance. Despite the success of deep co-training methods, their underlying mechanisms remain underexplored. This work revisits Cross Pseudo Supervision with dual heterogeneous backbones and introduces Knowledge Consultation (SegKC) to further enhance segmentation performance. The proposed SegKC achieves significant improvements on Pascal and Cityscapes benchmarks, with mIoU scores of 87.1\%, 89.2\%, and 89.8\% on Pascal VOC with the 1/4, 1/2, and full split partition, respectively, while maintaining a compact model architecture.
\end{abstract}

%% file: section/1_introduction.tex
\section{Introduction}

Semantic segmentation, a core task in deep learning and computer vision, assigns pixel-level labels to images \cite{atif2019reviewintro1, ghosh2019understandingreviewintro3, hao2020briefreviewintro2}. However, training these models requires extensive labeled data, which is costly and time-intensive; for example, annotating a single Cityscapes image with 19 classes takes approximately 1.5 hours \cite{yang2024unimatchv2consis1}. Semi-Supervised Learning tackles such phenomenon by using limited labeled data and incorporating unlabeled samples.

\begin{figure}[t]
    \centering
    \includegraphics[width=\linewidth]{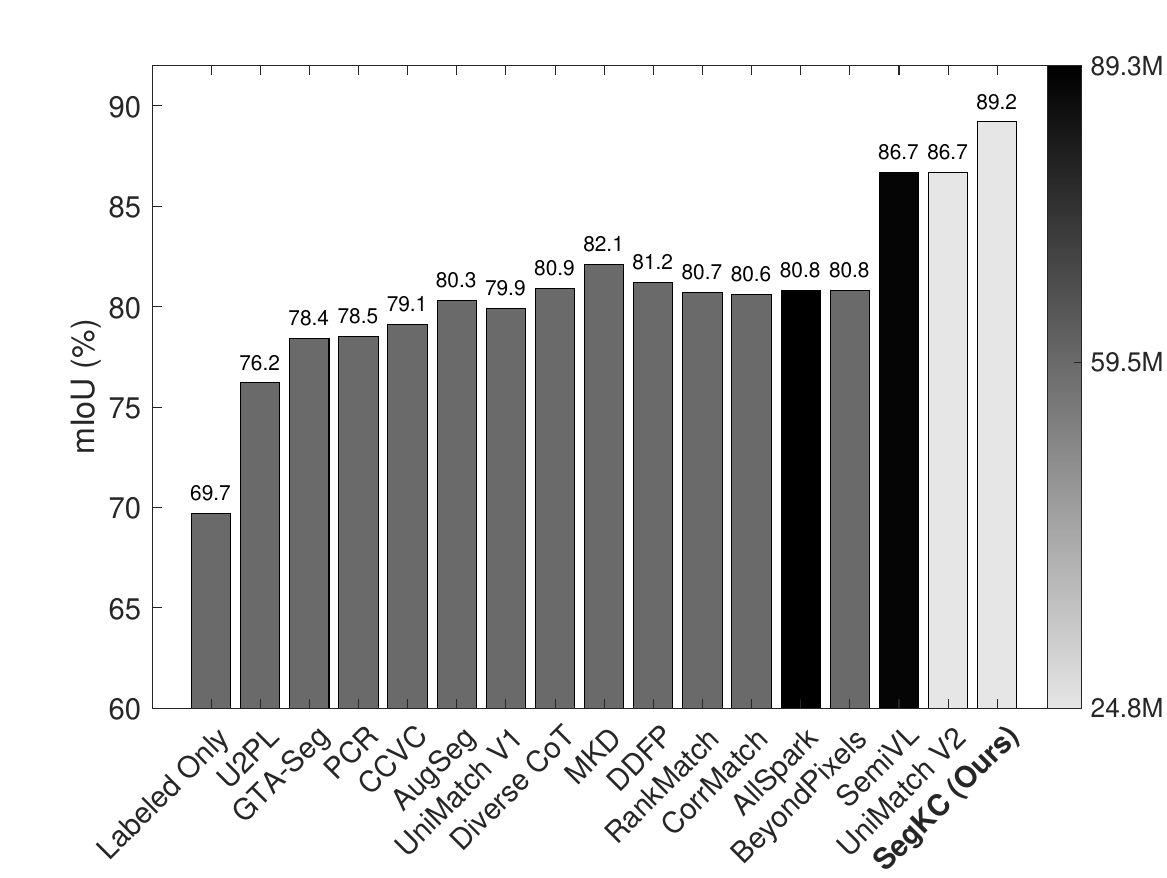}
    \caption{\textbf{Efficiency comparison of our method, SegKC, against state-of-the-art approaches on Pascal VOC original dataset}. We evaluate SegKC against leading methods using mean Intersection over Union (mIoU) and overall model size. With heterogeneous backbones in a Knowledge Consultation framework, SegKC offers a more dynamic knowledge transfer. While co-training both models increases costs, only the junior model is used for inference to maintain efficiency and practicality. This figure highlights SegKC’s strong performance on Pascal VOC original (1/2 partition), achieving 89.2\% mIoU with 24.8M parameters and setting a new benchmark for the dataset. Larger models are shown in darker colors to emphasize SegKC’s compact design and effectiveness. A detailed comparison is provided in Table~\ref{tab:pascal_origin}.}
    \label{fig:intro}
\end{figure}

Semi-Supervised Semantic Segmentation, a branch of Semi-Supervised Learning, improves segmentation accuracy with minimal labeled data by incorporating unlabeled images. A common approach is Pseudo-Labeling, where a model trained on labeled data generates pseudo-labels for unlabeled samples \cite{lee2013pseudo5, ouali2020pseudo17}. Current methods are broadly categorized into successive and parallel approaches. Successive methods, such as teacher-student frameworks, leverage an Exponential Moving Average (EMA) of the student’s weights to stabilize training and improve robustness. In contrast, parallel methods like CPS \cite{cui2023cpspseudo13} and CCVC \cite{wang2023ccvcpseudo15} optimize both models simultaneously to enhance the segmentation through collaborative learning.

Beyond these frameworks, Knowledge Distillation techniques offer another approach to refine Semi-Supervised Semantic Segmentation \cite{cho2019efficacyknowledgedissurvey2, gou2021knowledgedissurvey1}. Notably, Mutual Knowledge Distillation (MKD) \cite{yuan2023mkdkd1} combines weak-to-strong augmentations, EMA \cite{tarvainen2017meanteacher}, and cross-view training to boost performance. However, the integration of multiple components increases architectural complexity, which makes optimization more challenging. A fundamental limitation of cross-view methods like CPS \cite{cui2023cpspseudo13} and CCVC \cite{wang2023ccvcpseudo15} is their dependence on homogeneous backbones, which limits feature diversity at intermediate layers. To address this issue and diversify learned patterns, Diverse CoT \cite{li2023diversecot} introduces heterogeneous architectures. However, the need to train three models (3-CPS) significantly increases computational costs and creates practical constraints.

We refine cross-view co-training within a Knowledge Distillation framework, where a student model mimics a larger teacher to match performance efficiently. However, this unidirectional transfer of knowledge from the teacher model to the student model \cite{ke2019dualkddrawbacks2, kurata2020knowledgekddrawbacks1} limits the teacher from gaining insights captured by the student model. 

To overcome this limitation, we introduce the concept of Knowledge Consultation within the context of Semi-Supervised Semantic Segmentation. This framework adopts a senior-junior model instead of the conventional teacher-student relationship to enable bi-directional knowledge transfer.
Both models exchange knowledge to learn complementary features. Specifically, the junior model captures features overlooked by the senior model and integrates them into the senior model to improve the overall learning process. During inference, only the junior model is utilized. Our approach achieves cross-view co-training with Knowledge Consultation using two models, as opposed to the three required by 3-CPS. Despite the smaller size of the junior model, it outperforms state-of-the-art methods like CPS \cite{cui2023cpspseudo13}, CCVC \cite{wang2023ccvcpseudo15}, and Diverse CoT \cite{li2023diversecot}, which rely on combining outputs from multiple models for moderate performance. Comprehensive performance results are provided in Figure~\ref{fig:intro}.

Our contributions are summarized as follows:
\begin{itemize}[itemsep=0pt, topsep=0pt, partopsep=0pt, parsep=0pt]
    \item We introduce a novel approach of applying Cross Pseudo Supervision on heterogeneous backbones within a Knowledge Distillation framework, which uses only two models to enhance segmentation performance.
    \item We further propose a bi-directional Knowledge Consultation framework that replaces the teacher-student paradigm with a senior-junior structure. The new design enables knowledge to transfer at both feature and prediction levels.
    \item Our method attains a new state-of-the-art performance on Cityscapes and Pascal VOC 2012 datasets, including Pascal augmented and Pascal U$^2$PL split.
\end{itemize}

%% file: section/2_related_work.tex
\section{Related Work}

\begin{figure*}[!ht]
    \centering
    \includegraphics[width=\linewidth]{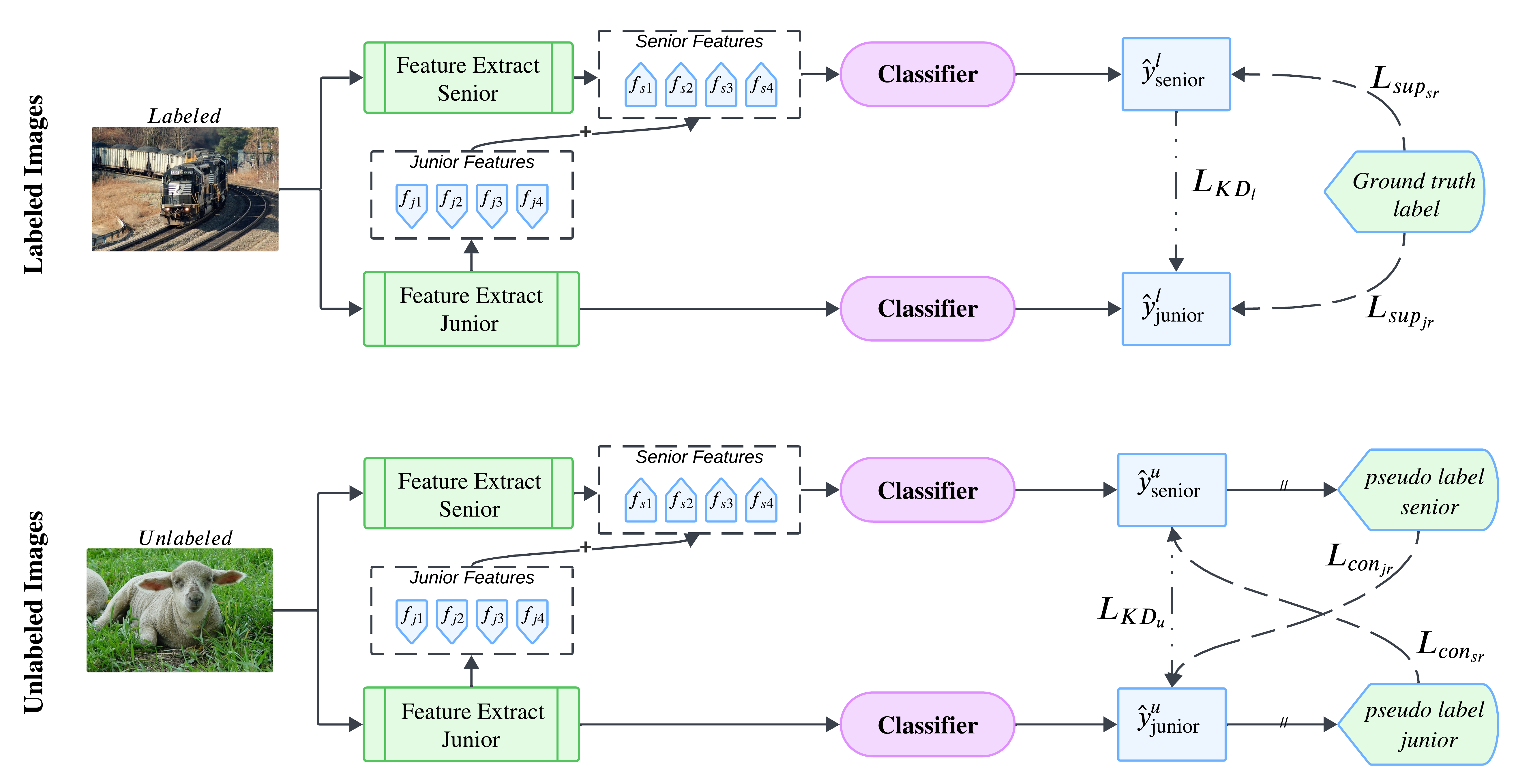}
    \caption{\textbf{Overview of the SegKC method} for Semi-Supervised Semantic Segmentation using labeled and unlabeled images. The method incorporates two distinct backbones, termed the senior and junior models. Knowledge transfers from the junior to the senior model to enrich the senior’s representations. The senior model distills knowledge back to the junior model and refines its understanding before generating final predictions. By using heterogeneous backbones and employing the junior model exclusively for inference, this design enhances Cross Pseudo Supervision while maintaining a compact and efficient architecture.}
    \label{fig:framework}
\end{figure*}

\subsection{Semi-Supervised Learning}
Semi-Supervised Learning is a technique that combines a small set of labeled data with a larger pool of unlabeled data to enhance model accuracy \cite{learning2006semissl3, van2020ssl2, zhou2021semissl1}. This method is particularly useful in tasks like image classification, where labeling extensive datasets is often impractical \cite{mehyadin2021classificationinfeasible1, peikari2018clusterinfeasible2}. There are two primary approaches to incorporating unlabeled data into training: Consistency Regularization and Pseudo-Labeling.

Consistency Regularization improves model accuracy and robustness by ensuring predictions remain consistent across variations of the same input. This approach is widely adopted due to its simplicity and effectiveness \cite{berthelot2019remixmatchconsis6, nguyen2024sequencematchconsis7, tan2023otmatchconsis10, wang2022freematchconsis9, zheng2022simmatchconsis8}. For example, MixMatch \cite{berthelot2019mixmatchconsis5} 
anticipates low-entropy labels for data-augmented unlabeled instances and blends both labeled and unlabeled data. 
Pseudo-Labeling \cite{grandvalet2004pseudo2, xie2020pseudo1, zoph2020pseudo3, pham2021pseudo4, lee2013pseudo5, kimhi2023pseudo6} involves training an initial model on labeled data. The model then uses its own predictions on unlabeled data as pseudo-labels for further training or for refining a second model \cite{zhang2021flexmatchpseudo7, huynh2022pseudo8}.  However, this approach can lead to confirmation bias. If the teacher model makes incorrect predictions, these errors may carry over to the student model during training phase \cite{mi2022activeconfirm1, li2023robustconfirm2, tang2021humbleconfirm3}. The Noisy Student method \cite{xie2020pseudo1} addresses this by adding noise to improve robustness of the model. 

Recent Semi-Supervised Learning approaches \cite{zhao2023rcpsbothmethods1, li2024openbothmethods2, li2024semibothmethods3, pieropan2022densebothmethods4, li2024studentbothmethods5} emphasize the benefits of combining both methods. An important example is FixMatch \cite{sohn2020fixmatchconsis3}, which uses weak-to-strong Consistency Regularization. It generates pseudo-labels from weakly augmented images and applies them to guide training on strongly augmented ones.

\subsection{Semi-Supervised Semantic Segmentation}
Semi-Supervised Semantic Segmentation leverages Semi-Supervised Learning to tackle the high cost of obtaining pixel-level annotations in computer vision tasks. This approach is particularly important in areas like autonomous driving and medical imaging, where detailed annotations demand significant resources \cite{taherdoost2024contribute1, kalluri2019contribute2, chebli2018contribute3, jiao2023contribute4, feyjie2020contribute5}. Key techniques in Semi-Supervised Semantic Segmentation that draw from Semi-Supervised Learning methods include Pseudo-Labeling, Consistency Regularization, and more recently, Knowledge Distillation.

In Pseudo-Labeling, methods such as CPS \cite{cui2023cpspseudo13} and CCVC \cite{wang2023ccvcpseudo15} use the first model’s predictions as pseudo-labels to train another and vice versa, thereby strengthening label connectivity and reducing confirmation bias. However, this approach does not fully eliminate confirmation bias as errors can still propagate through pseudo-labels. To address this issue, methods like ST++ \cite{yang2022st++pseudo16} apply adaptive re-weighting and selective Pseudo-Labeling. 

Consistency Regularization is another approach which improves model robustness by ensuring stable predictions across varied input perturbations. New frameworks such as FixMatch \cite{sohn2020fixmatchconsis3}, UniMatch \cite{yang2023unimatchv1consis2}, and UniMatch V2 \cite{yang2024unimatchv2consis1} incorporate methods like CutMix \cite{yun2019cutmix} and ClassMix \cite{olsson2021classmix} to apply weak-to-strong augmentations and guide model training. More advanced designs such as RankMatch \cite{mai2024rankmatchconsis4} and CorrMatch \cite{sun2024corrmatch} add pixel-wise and inter-pixel consistency to better capture segmentation boundaries.

Knowledge Distillation plays a crucial role in enhancing the performance of Semi-Supervised Semantic Segmentation \cite{li2023triplekd2, lu2024selfkd3, lu2024selfkd4, you2022simcvdkd5}. Methods like Mutual Knowledge Distillation (MKD) \cite{yuan2023mkdkd1} employ the teacher-student framework which enables the student model to learn from pseudo-labels produced by a more accurate teacher model. Other papers attempt Mean Teacher \cite{tarvainen2017meanteacher} 
to maintain the EMA teacher model as guidance. These methods reduce overfitting by using diverse augmented data to train the student model and ensure more reliable guidance, especially when labeled data is limited.

%% file: section/3_method.tex
\section{Method}
This section introduces our newly proposed Cross Pseudo Supervision on heterogeneous backbones, integrated within the Knowledge Consultation framework (SegKC).

\subsection{Problem statement}
In Semi-Supervised Semantic Segmentation, datasets include a small set of labeled samples and a significantly larger set of unlabeled samples. Let \( D_L \) represent the labeled dataset with instances \( \{(x_i^l, y_i)\}_{i=1}^{D_L} \), and \( D_U \) denote the unlabeled dataset with instances \( \{x_i^u\}_{i=1}^{D_U} \). Each input image \( x \) belongs to \( \mathbb{R}^{3 \times H \times W} \). The target segmentation is expressed as \( y \in \mathbb{R}^{K \times H \times W} \), with \( K \) being the number of predefined classes of the input image height of \( H \) and width of \( W \). 

Pseudo-Labeling methods first train a model on the labeled samples. This model then provides pseudo-labels \( \tilde{y} \) for the unlabeled data, which are used as additional supervision during subsequent training. Thus, the loss functions for the labeled and unlabeled data are defined as \( \mathcal{L}_l \) and \( \mathcal{L}_{u} \), respectively, as follows:

\begin{equation}
\mathcal{L}_{l} = \frac{1}{D_L} \sum_{i=1}^{D_L} \mathcal{L}_{\text{CE}}(\hat{y}_i^l, y_i)
\end{equation}
\begin{equation}
\mathcal{L}_{u} = \frac{1}{D_U} \sum_{i=1}^{D_U} \mathbbm{1}(\max(p_i) \geq \tau)\mathcal{L}_{\text{CE}}(\hat{y}_i^u, \tilde{y}_i^u)
\label{eq:unsupervised_loss}
\end{equation}
where \( \mathcal{L}_{\text{CE}} \) denotes the cross-entropy loss function. The two terms \( \hat{y}_i^l \) and \( \hat{y}_i^u \) represent the prediction of the labeled and unlabeled data, respectively. The \( y_i \) is the target label as defined, while \( \tilde{y}_i^u \) is the pseudo label applied for the unlabeled set. The term \( \mathbbm{1}(\max(p_i) \geq \tau) \) is a key component in FixMatch \cite{sohn2020fixmatchconsis3}, designed to mitigate the impact of noisy pseudo-labels. It incorporates a confidence threshold \( \tau \) to filter out unreliable pseudo-labels, ensuring that only labels exceeding this confidence level are used in training. This approach focuses on utilizing high-quality, manually labeled data in the early stages of training. As the model's confidence improves, pseudo-labeled images are gradually included in the process. Thus, the joint loss is a simple aggregation of labeled and unlabeled losses:

\begin{equation}
    \mathcal{L} = \mathcal{L}^l + \mathcal{L}^u
\end{equation}
Many existing methods refine components of the unsupervised loss term, while others enhance performance by making adjustments at the input or feature level of the model architecture. This study introduces significant modifications to both the model architecture and loss function. Figure~\ref{fig:framework} provides an overview of the SegKC framework.

\subsection{Heterogeneous Cross Pseudo Supervision}
Our SegKC approach leverages a co-training-based, two-branch network. The two sub-networks, known as the senior and junior models, are designed with distinct characteristics. Each sub-network consists of a feature extractor and a classifier.

To ensure the sub-networks generate accurate predictions, we deploy reference labels to supervise the training process for labeled data. This guides both sub-networks to produce semantically relevant outputs. The supervised loss is defined as: 
\begin{equation}
\mathcal{L}_{\text{sup}_i} = \frac{1}{D_L} \sum_{m=1}^{D_L} \frac{1}{H \times W} \sum_{n=0}^{H \times W} l_{\text{CE}}(\hat{y}_{mn}, y_{mn})
\end{equation}
The subscript \( i \) identifies whether the sub-networks correspond to the senior or junior model, while \( n \) refers to the \( n \)-th pixel in the \( m \)-th labeled image. The terms \( \hat{y}_{mn} \) and \( y_{mn} \) represent the predicted value and the target label for the \( n \)-th pixel in the \( m \)-th labeled image, respectively. Supervision using true labels is applied to both sub-networks to ensure meaningful predictions. The overall supervised loss is defined as \( \mathcal{L}_{\text{sup}} = 0.5 \times \left( \mathcal{L}_{\text{sup}_{sr}} + \mathcal{L}_{\text{sup}_{jr}} \right) \), where \( \mathcal{L}_{\text{sup}_{sr}} \) and \( \mathcal{L}_{\text{sup}_{jr}} \) correspond to the supervised losses for the senior and junior sub-networks.

For the unlabeled data, we apply a Pseudo-Labeling strategy that enables each sub-network to learn semantic information from the other. A prediction \( \hat{y}_{mn} \) generates a pseudo-label, defined as \( \tilde{y}_{mn} = \arg\max_c(\hat{y}_{mnc}) \), where \( \hat{y}_{mnc} \) denotes the \( c \)-th dimension of \( \hat{y}_{mn} \), and \( c \in \{1, \dots, K\} \) represents the category index. We then use the cross-entropy loss to optimize the model, and thus the consistency loss for each sub-network is expressed as follows:

\begin{equation}
\mathcal{L}_{\text{con}_{sr}} = \frac{1}{D_U} \sum_{m=1}^{D_U} \frac{1}{H \times W} \sum_{n=0}^{H \times W}{l}_{\text{CE}}\left(\hat{y}_{mn_{sr}}, \tilde{y}_{mn_{jr}}\right)
\end{equation}
\begin{equation}
\mathcal{L}_{\text{con}_{jr}} = \frac{1}{D_U} \sum_{m=1}^{D_U} \frac{1}{H \times W} \sum_{n=0}^{H \times W}{l}_{\text{CE}}\left(\hat{y}_{mn_{jr}}, \tilde{y}_{mn_{sr}}\right)
\end{equation}

where \( l_{CE} \) represents the cross-entropy loss function applied to the unlabeled dataset with respect to the confidence threshold, as outlined in Equation~\ref{eq:unsupervised_loss}. \( D_U \) denotes the total number of unlabeled images. Collectively, the cross-consistency loss for the unlabeled data is defined as \( \mathcal{L}_{\text{con}} = 0.5 \times \left( \mathcal{L}_{\text{con}_{sr}} + \mathcal{L}_{\text{con}_{jr}} \right) \), where \( \mathcal{L}_{\text{con}_{sr}} \) and \( \mathcal{L}_{\text{con}_{jr}} \) denote the consistency losses for the senior and junior sub-networks, respectively.

\subsection{Knowledge Consultation}
Our SegKC model operates on two layers: feature and prediction. At the feature level, representations from both the senior and junior backbones are extracted, with the junior model’s features incorporated into the senior model. The classifier then processes the combined features to produce prediction logits. This design enables bi-directional Knowledge Consultation, enhancing the learning capacity of the entire model architecture, as outlined in Figure~\ref{fig:framework}.

\input{table/pascal_original}

\input{table/pascal_augmented}

At the prediction layer, the softmax function with a temperature parameter \( \tau \) converts logits into probability distributions \( p(\mathbf{w}_m) \) and \( p(\mathbf{u}_m) \). Here, \( w_m \) represents the senior model’s output for the \( m \)-th image, while \( u_m \) corresponds to the junior model’s output for the same image. The \( c \)-th component of these vectors is computed as follows:
\begin{equation}
p(\mathbf{w}_m)^{(c)} = \frac{\exp(w_m^{(c)} / \tau)}{\sum_{d=1}^K \exp(w_m^{(d)} / \tau)}
\end{equation}
\begin{equation}
p(\mathbf{u}_m)^{(c)} = \frac{\exp(u_m^{(c)} / \tau)}{\sum_{d=1}^K \exp(u_m^{(d)} / \tau)}
\end{equation}
In this formulation, \( w_m^{(c)} \) and \( u_m^{(c)} \) correspond to the \( c \)-th components of the logits \( \mathbf{w}_m \) and \( \mathbf{u}_m \) respectively, where \( K \) represents the total number of classes. The subscript \( m \) denotes the \( m \)-th sample.

Knowledge Distillation is the process that enables \( p(\mathbf{u}_m)^{(c)} \) to mimic \( p(\mathbf{w}_m)^{(c)} \) across all classes and samples. This alignment is achieved by minimizing the Kullback-Leibler (KL) divergence, which is expressed as:
\begin{equation}
    \mathcal{L}_{\text{KD}} = \mathcal{L}_{\text{KL}}(p(\mathbf{w}_m) \| p(\mathbf{u}_m)) = \sum_{c=1}^C p(\mathbf{w}_m)^{(c)} \log \left( \frac{p(\mathbf{w}_m)^{(c)}}{p(\mathbf{u}_m)^{(c)}} \right)
\end{equation}

where \( p(\mathbf{w}_m)^{(c)} \) and \( p(\mathbf{u}_m)^{(c)} \) are probability distributions 
using a temperature-scaled softmax function. 
During model training, we aggregate the supervised loss \( \mathcal{L}_{\text{sup}} \), the consistency loss \( \mathcal{L}_{\text{con}} \), and the Knowledge Distillation loss \( \mathcal{L}_{\text{KD}} \) to form the total loss with weighting coefficients \( \lambda_1 \), \( \lambda_2 \), and \( \lambda_3 \) as follow:

\begin{equation}
    \mathcal{L} = \lambda_{1} \mathcal{L}_{\text{sup}} + \lambda_{2} \mathcal{L}_{\text{con}} + \lambda_{3} \mathcal{L}_{\text{KD}}
\end{equation}

%% file: table/pascal_original.tex
\begin{table*}[t]
\setlength\tabcolsep{1.7mm}
    \centering
    \begin{tabular}{lrcccccccc}
    \toprule
    
    Pascal original & Venue & Encoder & 1/16 (92) & 1/8 (183) & 1/4 (366) & 1/2 (732) & Full (1464) & Params \\
    
    \hhline{=========}

    Labeled Only & - & RN-101 & 45.1 & 55.3 & 64.8 & 69.7 & 73.5 & 59.5M \\

    U$^2$PL \cite{wang2022unriliablepseudo9} & CVPR'22 & RN-101 & 68.0 & 69.2 & 73.7 & 76.2 & 79.5 & 59.5M \\
		
    GTA-Seg \cite{jin2022sgtaseg} & NeurIPS'22 & RN-101 & 70.0 & 73.2 & 75.6 & 78.4 & 80.5 & 59.5M \\
    
    PCR \cite{xu2022pcr} & NeurIPS'22 & RN-101 & 70.1 & 74.7 & 77.2 & 78.5 & 80.7 & 59.5M \\

    CCVC \cite{wang2023ccvcpseudo15} & CVPR'23 & RN-101 & 70.2 & 74.4 & 77.4 & 79.1 & 80.5 & 59.5M \\
    
    AugSeg \cite{zhao2023augseg} & CVPR'23 & RN-101 & 71.1 & 75.5 & 78.8 & 80.3 & 81.4 & 59.5M \\
    
    UniMatch V1 \cite{yang2023unimatchv1consis2} & CVPR'23 & RN-101 & 75.2 & 77.2 & 78.8 & 79.9 & 81.2 & 59.5M \\

    Diverse CoT \cite{li2023diversecot}& ICCV'23 & RN-101 & 75.7 & 77.7 &  80.1 & 80.9 & 82.0 & 59.5M \\

    MKD \cite{yuan2023mkdkd1} & MM'23 & RN-101 & 76.1 & 77.8 & 80.4 & 82.1 & 83.8 & 59.5M \\

    DDFP \cite{wang2024towardsddfp}& CVPR'24 & RN-101 & 75.0 & 78.0 & 79.5 & 81.2 & 82.0 & 59.5M \\

    RankMatch \cite{mai2024rankmatchconsis4}& CVPR'24 & RN-101 & 75.5 & 77.6 & 79.8 & 80.7 & 82.2 & 59.5M \\
    
    CorrMatch \cite{sun2024corrmatch}& CVPR'24 & RN-101 & 76.4	& 78.5 & 79.4 & 80.6 & 81.8 & 59.5M \\

    AllSpark \cite{wang2024allspark} & CVPR'24 & MiT-B5 & 76.1 & 78.4 & 79.8 & 80.8 & 82.1 & 89.3M \\
    
    BeyondPixels \cite{howlader2024beyondpixel}& ECCV'24 & RN-101 & 77.3 & 78.6 & 79.8 & 80.8 & 81.7 & 59.5M \\
    
    SemiVL \cite{hoyer2025semivl} & ECCV'24 & CLIP-B & \textbf{84.0} & \textbf{85.6} & 86.0 & 86.7 & 87.3 & 88.0M \\
    
    UniMatch V2 \cite{yang2024unimatchv2consis1}& TPAMI'25 & DINOv2-S & 79.0 & 85.5 & 85.9 & 86.7 & 87.8 & \textbf{24.8M} \\

    \midrule

    SegKC & Ours & DINOv2-S & 82.0 & 84.4 & \textbf{87.1} & \textbf{89.2} & \textbf{89.8} & \textbf{24.8M} \\

    \bottomrule

    \end{tabular}
    \caption{\textbf{Comparison with state-of-the-art methods on the Pascal original dataset} under different partition protocols, with the DINOv2-Small (DINOv2-S) backbone setup and a crop size of 518.}
    \label{tab:pascal_origin}
    
\end{table*}

%% file: table/pascal_augmented.tex
\begin{table*}[t]
\setlength\tabcolsep{3.8mm}
    \centering
    \begin{tabular}{lrccccccc}
    \toprule
    
    Pascal augmented & Venue & Encoder & 1/16 (662) & 1/8 (1323) & 1/4 (2645) & Params \\
    
    \hhline{=======}

    Labeled Only & - & RN-101 & 67.5 & 71.1 & 74.2 & 59.5M \\
		
    GTA-Seg \cite{jin2022sgtaseg} & NeurIPS'22 & RN-101 & 77.8 & 80.5 & 80.6 & 59.5M \\
    
    CCVC \cite{wang2023ccvcpseudo15} & CVPR'23 & RN-101 & 77.2 & 78.4 & 79.0 & 119.0M \\
    
    AugSeg \cite{zhao2023augseg} & CVPR'23 & RN-101 & 77.1 & 77.3 & 78.8 & 59.5M \\
    
    UniMatch V1 \cite{yang2023unimatchv1consis2}& CVPR'23 & RN-101 & 78.1 & 78.9 & 79.2 & 59.5M \\

    MKD \cite{yuan2023mkdkd1} & MM'23 & RN-101 & 78.7 & 80.1 & 80.8 & 59.5M \\

    DDFP \cite{wang2024towardsddfp} & CVPR'24 & RN-101 & 78.3 & 78.9 & 79.8 & 59.5M \\

    RankMatch \cite{mai2024rankmatchconsis4}& CVPR'24 & RN-101 & 78.9 & 79.2 & 80.0 & 59.5M \\
    
    CorrMatch \cite{sun2024corrmatch}& CVPR'24 & RN-101 & 78.4	& 79.3 & 79.6 & 59.5M \\

    AllSpark \cite{wang2024allspark}& CVPR'24 & MiT-B5 & 78.3 & 80.0 & 80.4 & 89.3M \\
    
    BeyondPixels \cite{howlader2024beyondpixel} & ECCV'24 & RN-101 & 82.5 & 83.1 & 81.5 & 59.5M \\

    \midrule
    
    SegKC & Ours & DINOv2-S & \textbf{83.8} & \textbf{84.9} & \textbf{85.8} & \textbf{24.8M} \\

    \midrule

    U$^2$PL $^\dag$  \cite{wang2022unriliablepseudo9}& CVPR'22 & RN-101 & 77.2 & 79.0 & 79.3 & 59.5M \\

    GTA-Seg $^\dag$ \cite{jin2022sgtaseg}& NeurIPS'22 & RN-101 & 77.8 & 80.5 & 80.6 & 59.5M \\
			 
    PCR $^\dag$ \cite{xu2022pcr}& NeurIPS'22 & RN-101 & 78.6 & 80.7 & 80.8 & 59.5M \\

    CCVC $^\dag$ \cite{wang2023ccvcpseudo15}& CVPR'23 & RN-101 & 76.8 & 79.4 & 79.6 & 59.5M \\
    
    AugSeg $^\dag$ \cite{zhao2023augseg}& CVPR'23 & RN-101 & 79.3 & 81.5 & 80.5 & 59.5M \\
    
    UniMatch V1 $^\dag$ \cite{yang2023unimatchv1consis2}& CVPR'23 & RN-101 & 80.9 & 81.9 & 80.4 & 59.5M \\

    MKD $^\dag$ \cite{yuan2023mkdkd1}& MM'23 & RN-101 & 80.1 & 81.4 & 82.3 & 59.5M \\

    CorrMatch $^\dag$ \cite{sun2024corrmatch}& CVPR'24 & RN-101 & 81.3 & 81.9 & 80.9 & 59.5M \\

    AllSpark $^\dag$ \cite{wang2024allspark}& CVPR'24 & MiT-B5 & 81.6 & 82.0 & 80.9 & 89.3M \\
    
    \midrule
    
    SegKC $^\dag$ & Ours & DINOv2-S & \textbf{88.3} & \textbf{88.8} & \textbf{87.7} & \textbf{24.8M} \\

    \midrule

    \end{tabular}
    \caption{\textbf{Comparison with state-of-the-art methods on the Pascal augmented dataset} under different partition protocols, with the DINOv2-Small (DINOv2-S) backbone setup and a crop size of 518. The symbol $^\dag$ indicates the U$^2$PL partitioning scheme.}
    \label{tab:pascal_augmented}
    
\end{table*}

%% file: section/4_experiments.tex
\section{Experiments}
\subsection{Datasets}
\textbf{Pascal VOC 2012} \cite{everingham2015pascal} serves as a widely recognized benchmark used to assess the performance of semantic segmentation models. It contains 1,464 images with detailed semantic annotations together with 9,118 images featuring approximate masks across 21 categories. In our approach, we treat the finely annotated images as labeled data and use the remaining images as unlabeled. Additionally, we test our model on an augmented version of the dataset that includes extra annotations from the Segmentation Boundary Dataset (SBD) \cite{hariharan2011semanticsbd}. This expands the training set to a total of 10,582 images with augmented and U$^2$PL partitions. 

\textbf{Cityscapes}
\cite{cordts2016cityscapes} stands as another benchmark for Semi-Supervised Semantic Segmentation. It features urban street scenes with detailed pixel annotations across 19 classes. The dataset includes 2,975 images for training, 500 for validation, and 1,525 for testing. The combination of high-resolution images and intricate layouts makes it particularly well-suited for evaluating segmentation methods.

\subsection{Implementation details}
We use the basic DPT model \cite{ranftl2021visiondpt} for semantic segmentation. Many Semi-Supervised Semantic Segmentation methods \cite{li2023diversecot, wang2023ccvcpseudo15, yang2023unimatchv1consis2, yuan2023mkdkd1, zhao2023augseg} use ResNet as the backbone for DeepLabv3+. This backbone is typically pre-trained on ImageNet and fine-tuned with a low learning rate. However, ResNet encoders are becoming less effective. To improve feature representation and boost model performance, they are being replaced with ViT-based encoders, such as DINOv2 trained on large-scale datasets. This choice is motivated by the strong performance of DINOv2-Small, which surpasses many models despite having fewer parameters. 

The training resolution is set to 518 for Pascal and 738 for Cityscapes. For batch size, we use 16 (16 labeled and 16 unlabeled samples) for Pascal and 8 (8 labeled and 8 unlabeled samples) for Cityscapes. Experiments were conducted using 4 RTX A6000 GPUs for Pascal and 8 RTX A6000 GPUs for Cityscapes. We employ AdamW as the optimizer with a weight decay of 0.01. The learning rate for the encoder is set to \( 5e \)-6, while the decoder uses a rate that is 40 times higher. To adjust the learning rate, we apply a poly scheduler with the formula $\text{lr} \leftarrow \text{lr} \times \left( 1 - \frac{\text{iter}}{\text{total iter}} \right)^{0.9}$. 

The model undergoes training for 80 epochs on Pascal and 240 epochs on Cityscapes. During inference, we slightly adapt interpolation to ensure that both height and width of the images are divisible by 14. For Cityscapes, we follow previous approaches and use sliding window evaluation with a window size of 798. The mean Intersection-over-Union (mIoU) serves as the evaluation metric. Most importantly, we only report results from the junior branch as stated at the beginning of the paper. 

\subsection{Comparison with state-of-the-art methods}
\input{table/cityscapes}
On the \textbf{original Pascal VOC 2012 dataset}, Table~\ref{tab:pascal_origin} demonstrates robustness of our method across varying amounts of labeled data.  Using DINOv2-Small (24.8M parameters), we outperform state-of-the-art methods by 1.1\%, 2.5\%, and 2.0\% on splits with 366, 732, and 1,464 labeled samples, respectively.  Even with smaller splits of 92 and 183 labeled samples, our SegKC remains competitive at 82.0\% and 84.4\%, respectively, despite the inherent challenge of limited labeled data.

On the \textbf{augmented Pascal VOC 2012 dataset}, Table~\ref{tab:pascal_augmented} shows that our method significantly outperforms state-of-the-art. SegKC achieves gains of 1.3\%, 1.8\%, and 4.3\% over existing models on the 1/16, 1/8, and 1/4 labeled data splits, respectively.  Using the U$^2$PL split, we achieve 88.3\%, 88.8\%, and 87.7\%, exceeding previous best results by a substantial 6.7\%, 6.8\%, and 6.8\%. This result sets a new record for the benchmark.

On the \textbf{Cityscapes dataset}, our method significantly outperforms the recent state-of-the-art UniMatch V2.  Table~\ref{tab:cityscapes} shows SegKC achieves accuracies of 81.2\%, 82.4\%, 83.4\%, and 83.6\% on the 186, 372, 744, and 1488 labeled data splits, respectively. The results of SegKC surpasses UniMatch V2 in all cases.

\subsection{Ablation study}

\input{table/ablation_components}

\textbf{More components, greater performance.}
Table~\ref{tab:components} demonstrates that the inclusion of all components consistently delivers the best performance across all datasets and splits, with larger differences observed in smaller sample sizes. On the Pascal original dataset with a 1/16 split, the full combination achieves an accuracy of 82.0\%, a significant improvement of 36.9\% compared to the configuration with only the supervised loss (45.1\%) and 7.9\% higher than the setup excluding the knowledge distillation loss (74.1\%). Similar trends are evident on the Pascal augmented dataset, where the full setup achieves 83.8\% and 88.3\% on the augmented and U$^2$PL 1/16 split, which outperforms configurations without the knowledge distillation loss or with only the supervised loss by 0.6\%, 0.5\%, 16.3\%, and 20.5\%, respectively. On the Cityscapes dataset with a 1/16 split, the full setup attains 81.2\%, which is 14.9\% higher than the configuration with only the supervised loss (66.3\%).

\textbf{Heterogeneous outperforms homogeneous in Cross Pseudo Supervision.} Table~\ref{tab:hetero_homo} presents an ablation study to test whether using different branches is more effective than using identical ones, as suggested in CPS and CCVC. In our training setup, we use DINOv2-Base for the senior model and DINOv2-Small for the junior model. For the counterpart configuration, both branches use DINOv2-Small as their encoder backbones to form a peer setup. The heterogeneous backbone demonstrates superior performance in all three datasets including Pascal original, Pascal augmented, and Cityscapes. 
On the one hand, the new design achieves 82.0\% while the homogeneous backbone scores 68.7\% on the  1/16  split of the Pascal original dataset. This is indeed the largest difference in all tests, which might be attributed to abnormalities in the data distribution.  Similarly, on the Pascal augmented dataset, the heterogeneous backbone scores 85.8\% and 88.3\% on the 1/4 and U$^2$PL 1/16 split. The results of the new model outperform the homogeneous model’s 84.6\% and 85.1\%. On Cityscapes, the heterogeneous setup achieves 81.2\% on the 1/16 split and surpasses the homogeneous configuration at 80.0\%.

\input{table/ablation_peers_pascal}

\textbf{Larger senior leads to a better junior's performance.}
A more powerful senior model intuitively provides richer features, enhancing the junior model’s performance. This trend is apparent with DINOv2 backbones, as shown in Table~\ref{tab:backbones}: the large senior and small junior (SegKC-L-S) configuration consistently outperforms the base senior and small junior (SegKC-B-S) setup across all splits. On the 1/16 Pascal original split, the larger senior model achieves a score of 79.6\%, surpassing the base variant by 8.8\%. A similar trend is observed in Pascal augmented, where the 1/8 split records 84.8\%, exceeding the base setup by 0.5\%. Under U$^2$PL split, SegKC-L-S consistently maintains an advantage, scoring 84.9\% on the 1/4 split, which exceeds SegKC-B-S by 1.1\%. To accommodate our computational constraints, this experiment was conducted on 8 RTX A6000 GPUs with a batch size of 8 (8 labeled and 8 unlabeled) instead of 16 (16 labeled and 16 unlabeled), as used in our other Pascal dataset experiments. 

\input{table/ablation_large_senior}

\subsection{Qualitative Comparison}
\begin{figure}[!ht]
    \centering
    \includegraphics[trim={0.8cm 7.2cm 0.75cm 0}, clip, width=\linewidth]{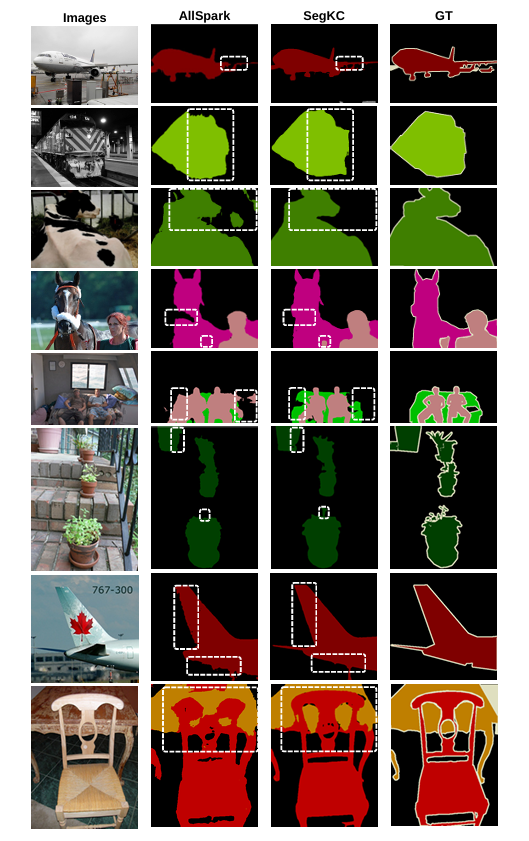}
    \caption{\textbf{Qualitative comparison on the original Pascal VOC 2012} dataset between AllSpark, SegKC (ours), and ground truth (GT) using the 1/2 partition.
    }
    \label{fig:qualitative_pascal}
\end{figure}

\begin{figure}[!ht]
    \centering
    \includegraphics[trim={0.2cm 11.3cm 0.2cm 0}, clip,width=\linewidth]{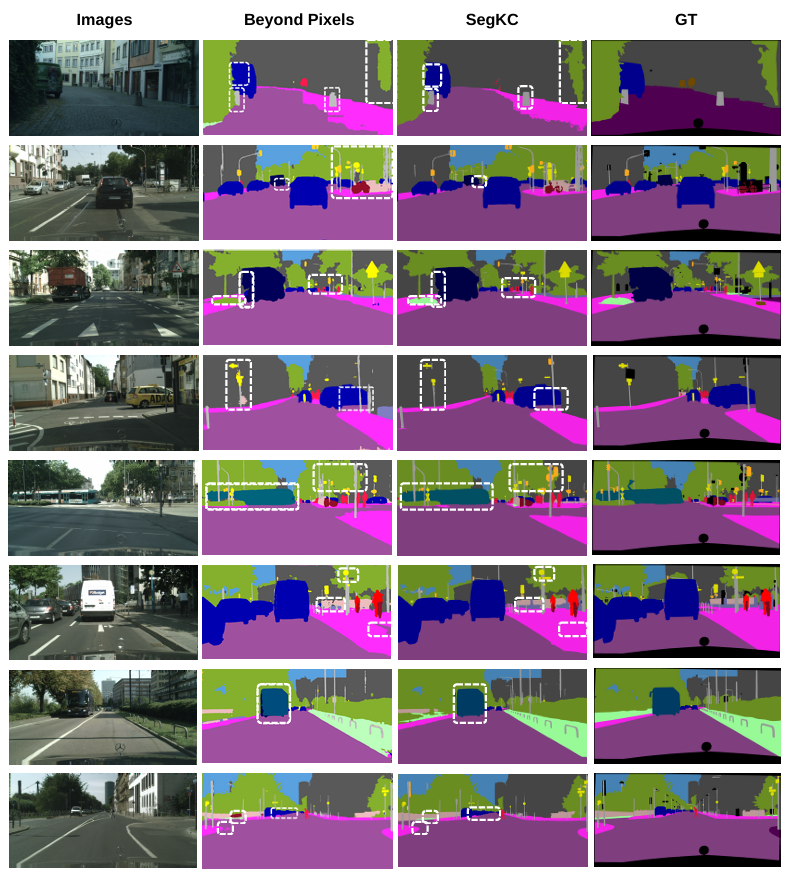}
    \caption{\textbf{Qualitative comparison on the Cityscapes} dataset between BeyondPixels, SegKC (ours), and ground truth (GT) using the 1/2 partition.
    }
    \label{fig:qualitative_cityscapes}
\end{figure}

Figures~\ref{fig:qualitative_pascal} and~\ref{fig:qualitative_cityscapes} illustrate the segmentation performance of SegKC across images. In Figure~\ref{fig:qualitative_pascal}, AllSpark \cite{wang2024allspark}, a state-of-the-art method, produces less accurate segmentation details compared to SegKC. SegKC’s ability to generate sharper boundaries and resolve intricate image features highlights its superior performance. Figure~\ref{fig:qualitative_cityscapes} presents results on urban street scenes, where SegKC consistently outperforms Beyond Pixels \cite{howlader2024beyondpixel}. It achieves precise segmentation while maintaining a high level of accuracy compared to the ground truth. These results demonstrate SegKC’s robustness and reliability across diverse scenarios.

%% file: table/cityscapes.tex
\begin{table*}[t]
\setlength\tabcolsep{3.0mm}
    \centering
    \begin{tabular}{lrccccccc}
    \toprule
    
    Cityscapes & Venue & Encoder & 1/16 (186) & 1/8 (372) & 1/4 (744) & 1/2 (1488) & Params \\
    
    \hhline{========}

    Labeled Only & - & RN-101 & 66.3 & 72.8 & 75.0 & 78.0 & 59.5M \\
	
    U$^2$PL \cite{wang2022unriliablepseudo9} & CVPR'22 & RN-101 & 74.9 & 76.5 & 78.5 & 79.1 & 59.5M \\
		
    GTA-Seg \cite{jin2022sgtaseg}& NeurIPS'22 & RN-101 & 69.4 & 72.0 & 76.1 & - & 59.5M \\
    

    CCVC \cite{wang2023ccvcpseudo15}& CVPR'23 & RN-101 & 74.9 & 76.4 & 77.3 & - & 119.0M \\
    
    AugSeg \cite{zhao2023augseg} & CVPR'23 & RN-101 & 75.2 & 77.8 & 79.6 & 80.4 & 59.5M \\
    
    UniMatch V1 \cite{yang2023unimatchv1consis2}& CVPR'23 & RN-101 & 76.6 & 77.9 & 79.2 & 79.5 & 59.5M \\

    Diverse CoT \cite{li2023diversecot} & ICCV'23 & RN-101 & 75.7 & 77.4 &  78.5 & - & 59.5M \\

    MKD \cite{yuan2023mkdkd1}& MM'23 & RN-101 & 77.2 & 79.2 & 80.8 & 81.0 & 59.5M \\

    DDFP \cite{wang2024towardsddfp}& CVPR'24 & RN-101 & 77.1 & 78.2 & 79.9 & 80.8 & 59.5M \\

    RankMatch \cite{mai2024rankmatchconsis4}& CVPR'24 & RN-101 & 77.1 & 78.6 & 80.0 & 80.7 & 59.5M \\
    
    CorrMatch \cite{sun2024corrmatch}& CVPR'24 & RN-101 & 77.3	& 78.5 & 79.4 & 80.4 & 59.5M \\

    AllSpark \cite{wang2024allspark}& CVPR'24 & MiT-B5 & 78.3 & 79.2 & 80.6 & 81.4 & 89.3M \\
    
    BeyondPixels \cite{howlader2024beyondpixel} & ECCV'24 & RN-101 & 78.5 & 79.2 & 80.9 & 81.3 & 59.5M \\
    
    SemiVL \cite{hoyer2025semivl}& ECCV'24 & CLIP-B & 77.9 & 79.4 & 80.3 & 80.6 & 59.5M \\
    
    UniMatch V2 \cite{yang2024unimatchv2consis1} & TPAMI'25 & DINOv2-S & 80.6 & 81.9 & 82.4 & 82.6 & \textbf{24.8M} \\

    \midrule

    SegKC & Ours & DINOv2-S & \textbf{81.2} & \textbf{82.4} & \textbf{83.4} & \textbf{83.6} & \textbf{24.8M} \\

    \midrule

    \end{tabular}
    \caption{\textbf{Comparison with state-of-the-art methods on the Cityscapes dataset} under different partition protocols, with the DINOv2-Small (DINOv2-S) backbone setup and a crop size of 712.}
    \label{tab:cityscapes}
    
\end{table*}

%% file: table/ablation_components.tex
\begin{table*}[t]\centering
\vspace{-10pt}
\setlength\tabcolsep{1.6mm}
\begin{tabular}{ccc|ccccc|ccc|ccc|cccc}
\hline
\multicolumn{3}{c}{Module} & \multicolumn{5}{c}{Pascal original} & \multicolumn{6}{c}{Pascal augmented} & \multicolumn{4}{c}{Cityscapes} \\

\hline

$\mathcal{L}_{sup}$ & $\mathcal{L}_{con}$ & $\mathcal{L}_{KD}$ & 1/16 & 1/8 & 1/4 & 1/2 & Full & 1/16 & 1/8 & 1/4 & 1/16 $^\dag$ & 1/8 $^\dag$ & 1/4 $^\dag$ & 1/16 & 1/8 & 1/4 & 1/2 \\

\hline

\checkmark & & & 45.1 & 55.3 & 64.8 & 69.7 & 73.5 & 67.5 & 71.1 & 74.2 & 67.8 & 71.6 & 75.8 & 66.3 & 72.8 & 75.0 & 78.0 \\
\checkmark & \checkmark & & 74.1 & 80.3 & 85.8 & 86.4 & 88.3 & 83.2 & 84.6 & 84.8 & 87.8 & 88.2 & 87.1 & 81.0 & 82.0 & 82.9 & 83.4 \\
\checkmark & \checkmark & \checkmark & \textbf{82.0} & \textbf{84.4} & \textbf{87.1} & \textbf{89.2} & \textbf{89.9} & \textbf{83.8} & \textbf{84.9} & \textbf{85.8} & \textbf{88.3} & \textbf{88.8} & \textbf{87.7} & \textbf{81.2} & \textbf{82.4} & \textbf{83.4} & \textbf{83.6} \\

\hline

\end{tabular}
\caption{\textbf{Ablation study on the effectiveness of individual modules} in our method across Pascal original, Pascal augmented, and Cityscapes datasets. With the incorporation of the supervised loss ($\mathcal{L}_{sup}$), consistency loss ($\mathcal{L}_{con}$), and Knowledge Distillation loss ($\mathcal{L}_{KD}$), the model achieves the best performance across all splits. This demonstrates their complementary contributions toward improving model accuracy. $\dag$ denotes the same split as U$^2$PL.}
\label{tab:components}
\end{table*}

%% file: table/ablation_peers_pascal.tex
\begin{table*}[t]
\setlength\tabcolsep{1.8mm}
    \centering
    \begin{tabular}{cccccc|ccc|ccc|cccc}
    \toprule
    
    \multirow{2}{*}{Model} & \multicolumn{5}{c}{Pascal original} & \multicolumn{3}{c}{Pascal augmented}
    & \multicolumn{3}{c}{Pascal U$^2$PL split}
    & \multicolumn{4}{c}{Cityscapes} \\
    
    \cmidrule(lr){2-6}\cmidrule(lr){7-9}\cmidrule(lr){10-12}\cmidrule(lr){13-16}
    
    & 1/16 & 1/8 & 1/4 & 1/2 & Full & 1/16 & 1/8 & 1/4 & 1/16 $^\dag$ & 1/8 $^\dag$ & 1/4 $^\dag$ & 1/16 & 1/8 & 1/4 & 1/2 \\
    
    \midrule
    
    SegKC-S-S & 68.7 & 81.6 & 86.4 & 87.6 & 88.4 & 83.3 & 84.4 & 84.6 & 85.1 & 87.3 & 86.0 & 80.0 & 81.6 & 82.8 & 82.9 \\

    SegKC-B-S & \textbf{82.0} & \textbf{84.4} & \textbf{87.1} & \textbf{89.2} & \textbf{89.9} & \textbf{83.8} & \textbf{84.9} & \textbf{85.8} & \textbf{88.3} & \textbf{88.8} & \textbf{87.7} & \textbf{81.2} & \textbf{82.4} & \textbf{83.4} & \textbf{83.6} \\

    \bottomrule
    
    \end{tabular}
    \caption{\textbf{Ablation study on the effectiveness of heterogeneous backbones}. The results show that the heterogeneous backbones (SegKC-\textbf{B}-\textbf{S} refers to the DINOv2-Base model for the senior branch and the DINOv2-Small model for the junior branch) consistently outperform the homogeneous backbones (SegKC-\textbf{S}-\textbf{S} refers to the DINOv2-Small model for both the senior and junior branches) across all splits of the Pascal original, Pascal augmented, and Cityscapes dataset. The symbol $\dag$ indicates the same split as U$^2$PL.}
    \label{tab:hetero_homo}
\end{table*}

%% file: table/ablation_large_senior.tex
\begin{table*}[t]
\setlength\tabcolsep{1.5mm}
    \centering
    \begin{tabular}{ccccccc|ccc|ccc}
    \toprule
    
    \multirow{2}{*}{Model} & \multirow{2}{*}{Batch size} & \multicolumn{5}{c}{Pascal original} & \multicolumn{3}{c}{Pascal augmented} & \multicolumn{3}{c}{Pascal U$^2$PL split} \\
    
    \cmidrule(lr){3-7}\cmidrule(lr){8-10}\cmidrule(lr){11-13}
    
    & & 1/16 & 1/8 & 1/4 & 1/2 & Full & 1/16 & 1/8 & 1/4 & 1/16$^\dag$ & 1/8$^\dag$ & 1/4$^\dag$ \\
    
    \midrule
    
    SegKC-B-S & 8 & 70.8 & 79.4 & 83.9 & 87.2 & 88.0 & 83.2 & 84.3 & 85.1 & 85.8 & 86.1 & 83.8 \\
    SegKC-L-S & 8 & \textbf{79.6} & \textbf{81.1} & \textbf{84.5} & \textbf{87.6} & \textbf{88.7} & \textbf{83.4} & \textbf{84.8} & \textbf{85.4} & \textbf{86.5} & \textbf{86.7} & \textbf{84.9} \\

    \bottomrule
    
    \end{tabular}
    \caption{\textbf{Ablation study on the impact of senior model capacity}. The results compare the performance of using DINOv2-\textbf{B}ase (in SegKC-\textbf{B}-S model) and DINOv2-\textbf{L}arge (in SegKC-\textbf{L}-S model) as the senior model while keeping DINOv2-\textbf{S}mall (in SegKC-B-\textbf{S} model and SegKC-L-\textbf{S} model) as the junior model. Using a batch size of 8 (8 labeled and 8 unlabeled), the larger senior consistently shows dominant results. This demonstrates that increased computational resources will yield even stronger results.}
    \label{tab:backbones}
\end{table*}

%% file: section/5_conclusion.tex
\section{Conclusion}
This paper introduces SegKC, a novel method that enhances Semi-Supervised Semantic Segmentation by leveraging Cross Pseudo Supervision on heterogeneous backbones. Our method adopts a bi-directional Knowledge Consultation, which shifts from the traditional teacher-student structure to a senior-junior model. Such design transfers knowledge in both directions at the feature and prediction levels. The benchmark results of various datasets illustrate that SegKC achieves superior performance and demonstrates its computational efficiency. We believe its generality enables applications beyond semantic segmentation, including self-supervised learning and efficiency improvements for large-scale deployment.